\documentclass[11pt]{article}
\usepackage{acl2013}
\usepackage[colorlinks,allcolors=blue]{hyperref}

\usepackage{float}
\usepackage{graphicx}
\usepackage[hyphenbreaks]{breakurl}
\floatstyle{ruled}
\newfloat{program}{thp}{lop}
\floatname{program}{Program}
\usepackage{times}
\usepackage{url}
\usepackage{latexsym}
\usepackage{placeins}
\usepackage{footnotehyper}
\usepackage{float}
\usepackage{breakcites}
\usepackage{amsmath}
\usepackage{tabularx,booktabs,caption}% http://ctan.org/pkg/{tabularx,booktabs,caption}
\makeatletter
\renewcommand\paragraph{\@startsection{paragraph}{4}{\z@}%
                                    {3.25ex \@plus1ex \@minus.2ex}%
                                    {-1em}%
                                    {\normalfont\normalsize}}
\makeatother

\title{Using Statistical and Semantic Models\\ for Multi-Document Summarization}

\author{Divyanshu Daiya \thanks{\hspace{0.1cm} 	Both the authors have contributed equally to this work.}\\
  LNM Institute of Information \\
  Technology\\
  Jaipur, Rajasthan 302031 \\
  {\small{daiyadivyanshu@gmail.com}} \\\And
  Anukarsh Singh \footnotemark[1]\\
  LNM Institute of Information \\
  Technology\\
  Jaipur, Rajasthan 302031 \\
  {\small{anukarshsingh1@gmail.com}} \\\And
  Mukesh Jadon \\
  LNM Institute of Information \\
  Technology\\
  Jaipur, Rajasthan 302031 \\
  {\small{jadonmukesh30@gmail.com}} \\}

\date{}

\begin{document}
\maketitle
\begin{abstract}
 We report a series of experiments with different semantic models on top of various statistical models for extractive text summarization. Though statistical models may better capture word co-occurrences and distribution around the text, they fail to detect the context and the sense of sentences /words as a whole. Semantic models help us gain better insight into the context of sentences. We show that how tuning weights between different models can help us achieve significant results on various benchmarks. Learning pre-trained vectors used in semantic models further, on given corpus, can give addition spike in performance. Using weighing techniques in between various statistical models too further refines our result. For Statistical models, we have used TF/IDF, TextRAnk, Jaccard/Cosine Similarities. For Semantic Models, we have used WordNet-based Model and proposed two models based on Glove Vectors and Facebook's InferSent. We tested our approach on DUC 2004 dataset, generating  100-word summaries. We have discussed the system, algorithms, analysis and also proposed and tested possible improvements. ROUGE scores \cite{lin2004rouge} were used to compare to other summarizers.
  
\end{abstract}

\section{Introduction}

Automatic Text Summarization deals with the task of condensing documents into a summary, whose level is similar to a human-generated summary. It is mostly distributed into two distinct domains, i.e., Abstractive Summarization and Extractive Summarization. Abstractive summarization( Dejong et al. ,1978)  involves models to deduce the crux of the document. It then presents a summary consisting of words and phrases that were not there in the actual document, sometimes even paraphrasing\cite{DBLP:journals/corr/RocktaschelGHKB15}. A state of art method proposed by  Wenyuan Zeng \cite{DBLP:journals/corr/ZengLFU16} produces such summaries with length restricted to 75. There have been many recent developments that produce optimal results, but it is still in a developing phase. It highly relies on natural language processing techniques, which is still evolving to match human standards. These shortcomings make abstractive summarization highly domain selective. As a result, their application is skewed to the areas where NLP techniques have been superlative. Extractive Summarization, on the other hand, uses different methods to identify the most informative/dominant sentences through the text, and then present the results, ranking them accordingly.
In this paper, we have proposed two novel stand-alone summarization methods.The first method is  based on Glove Model  \cite{pennington2014glove},and other is based on Facebook's InferSent  \cite{conneau2017supervised}. We have also discussed how we can effectively subdue shortcomings of one model by using it in coalition with models which capture the view that other faintly held. 

\section{Related Work}
 A vast number of methods have been used for document summarization. Some of the methods include determining the length and positioning of sentences in the text \cite{radev2004centroid}, deducing centroid terms to find the importance of text  \cite{radev2004centroid} and setting a threshold on average TF-IDF scores.  Bag-of-words approach, i.e., making sentence/Word freq matrix, using a signature set of words and assigning them weights to use them as a criterion for importance measure  \cite{lin2000automated} have also been used. Summarization using weights on high-frequency words  \cite{nenkova2006compositional} describes that high-frequency terms can be used to deduce the core of document.\\
While semantic summarizers like Lexical similarity is based on the assumption that important sentences are identified by strong chains \cite{gupta2011summarizing,barrera2012combining,murdock2006aspects}. In other words, it relates sentences that employ words with the same meaning (synonyms) or other semantic relation. It uses WordNet \cite{miller1990introduction} to find similarity among words that apply to Word Frequency algorithm.POS(Part of Speech) Tagging and  WSD(Word Sense Disambiguation) are common among semantic summarizers. 
Graphical summarizers like TextRank have also provided great benchmark results.TextRank assigns weights to important keywords from the document using graph-based model and sentences which capture most of those concepts/keywords are ranked higher) \cite{barrera2012combining,mihalcea2004textrank} TextRank uses Google's PageRank (Brin and Page, 1998) for graphical modeling. Though semantic and graphical models may better capture the sense of document but miss out on statistical view.\\
There is a void of hybrid summarizers; there haven't been many studies made in the area.Wong\cite{wong2008extractive} conducted some preliminary research but there isn't much there on benchmark tests to our knowledge.
We use a mixture of statistical and semantic models, assign weights among them by training on field-specific corpora. As there is a significant variation in choices among different fields. We support our proposal with expectations that shortcomings posed by one model can be filled with positives from others. We deploy experimental analysis to test our proposition. 

\section{Proposed Approach}

For Statistical analysis we use Similarity matrices, word co-occurrence/ n-gram model, andTF/IDF matrix. For semantic analysis we use custom Glove based model, WordNet based Model and Facebook InferSent \cite{conneau2017supervised} based Model. For Multi-Document Summarization,after training on corpus, we assign weights among the different techniques .We store the sense vector for documents, along with weights, for future reference. For Single document summarization, firstly we calculate the sense vector for that document and calculate the nearest vector from the stored Vectors, we use the weights of the nearest vector. We will describe the flow for semantic and statistical models separately.

\subsection{Prepossessing}
We discuss, in detail, the steps that are common for both statistical and semantic models.

%\url{http://www.acl2013.org}. A Microsoft Word
\subsubsection{Sentence Tokenizer}
We use NLTK sentence tokenizer sent\_tokenize(), based on PUNKT tokenizer, pre-trained on a corpus. It can differentiate between Mr. , Mrs. and other abbreviations etc. and the normal sentence boundaries.\cite{kiss2006unsupervised}\\
Given a document $D$ we tokenize it into sentences as {\textless$\bf s_1 ,s_2,s_3,s_4 ... s_n $\textgreater}.

\subsubsection{Cleaning}
Replacing all the special characters with spaces for easier word-tagging and Tokenizing.

\subsubsection{Word Tokenizer}
We use NLTK word tokenizer, which is a Penn
Treebank–style tokenizer, to tokenize words.We calculate the total unique words in the Document. If we can write any sentence as:-\\
\linebreak
 $s_i \rightarrow$ {\textless$\bf w_I,w_J,w_K,w_L,..$ \textgreater}, ${\tt i} \in (1,n)$ \\
 \linebreak
 Then the number of unique words can be represented as:- \\
\[ {\tt (I,J,K,L....)} \subset (1,.. M)\]  
\[n\rightarrow Total sentences ,M\rightarrow  Total unique words\] 

\subsection{Using Stastical Models}
\label{sect:pdf}

\subsubsection{Similarity/Correlation Matrices}

\paragraph{\textbf{Frequency Matrix generation}:}
Our tokenized words contain redundancy due to digits and transitional words such as ``and'', ``but'' etc., which carry little information. Such words are termed stop words.\cite{wilbur1992automatic} We removed stop words and words occurring in \textless 0.2\% and \textgreater 15\% of the documents (considering the word frequency over all documents). After the removal, the no. of unique words left in the particular document be {\tt p} where {\tt p\textless m} (where {\tt m} is the total no. of unique words in our tokenized list originally). We now formulate a matrix  $F_{\tt n \times p}$ where {\tt n} is the total number of sentences and {\tt p} is the total number of unique words left in the document. Element $e_{ij}$ in the matrix $F_{\tt n \times p}$ denotes frequency of $j^{th}$ unique word in the $i^{th}$ sentence.

\paragraph{\textbf{Similarity/Correlation Matrix generation}:}
We now have have sentence word frequency vector $\bf{Sf_{i}}$ as {\textless$\bf f_{i1},f_{i2},f_{i3},... f_{i4} $\textgreater} where $f_{ia}$ denotes frequency of $a^{th}$ unique word in the $i^{th}$ sentence. We now compute,
        \[ {\tt Sentence\_similarity ( Sf_{i},Sf_{j}) } \]
We use two similarity measures :
\begin{enumerate}
  \item Jaccard Similarity
  \item Cosine Similarity
%  \item Minkowski Distance
\end{enumerate}

We generate the similarity matrix $Sim^{j}_{n\times  n}$ for each of the similarity Measure, where ${\tt j}$ indexes the similarity Measure. Element $E_{ij}$ of $Sim^{j}_{n\times  n}$ denotes similarity between ${\tt i^{th}}$ and ${\tt j^{th}}$ sentence. Consequentially, we will end up with $Sim^{1}_{n\times  n}$ and $Sim^{2}_{n\times  n}$, corresponding to each similarity measure.

\subparagraph{Jaccard Similarity:}
For some sets A and B, \textless a,b,c,... \textgreater  and \textless x,y,z,... \textgreater  respectively, the Jaccard Similarity is defined as:-
\[ {\tt Jaccard\_similarity (A,B) } \leftarrow \frac{n(A\cap B)}{n(A\cup B)}\]

\subparagraph{Cosine Similarity:}
The Cosine distance between `u' and `v', is defined as:-
    
       \[{\tt Cosine\_similarity (A,B) } \leftarrow  1 - \frac{u \cdot v}{||u|| ||v||}\]
    
    where $`u \cdot v`$ is the dot product of $`u`$ and $`v`$.

\subsubsection{PageRank}

PageRank algorithm \cite{page1999pagerank}, devised to rank web pages, forms the core of Google Search. It roughly works by ranking pages according to the number and quality of outsourcing links from the page. For NLP, a PageRank based technique ,TextRank has been a major breakthrough in the field. TextRank based summarization has seeded exemplary results on benchmarks. We use a naive TextRank analogous for our task.\\
Given $n$ sentences {\textless $\bf s_1 ,s_2,s_3,.. s_n $\textgreater}, we intend to generate PageRank or probability distribution matrix $\bf{R_{n\times 1}}$, \[\begin{bmatrix}
    Pr(s_1) \\
    Pr(s_2) \\
    \vdots \\      
    Pr(s_n)
\end{bmatrix}\],
where $Pr(s_k)$ in original paper denoted probability with which a randomly browsing user lands on a particular page. For the summarization task, they denote how strongly a sentence is connected with rest of document, or how well sentence captures multiple views/concepts. The steps are as:
\begin{enumerate}
  \item Initialize $\bf{R}$ as,\\
      \[
    \begin{bmatrix}
        Pr(s_1) \\
        Pr(s_2) \\
        \vdots \\      
        Pr(s_n)
    \end{bmatrix}
    =
    \begin{bmatrix}
        \frac{1}{n} \\
        \frac{1}{n} \\
        \vdots \\      
        \frac{1}{n}
    \end{bmatrix}
    \]
  \item Define $\bf{d}$, probability that randomly chosen sentence is in summary and $\bf{\varepsilon}$ as measure of change i.e. to stop computation when difference between to successive $\bf{R}$ computations recedes below $\varepsilon$.
     
  \item Using cosine-similarity matrix $Sim^{2}_{n\times  n}$, we generate the following equation as a measure for relation between sentences:-
  
    \[
        \bf{R} =
        \begin{bmatrix}
           (1-d)/n \\
            (1-d)/n \\
            \vdots \\      
            (1-d)/n \\
        \end{bmatrix}
        +
        d\times Sim^{2}_{n\times  n}\times \bf{R}
   %     \begin{bmatrix}
    %        sim(s_1,s_1) & sim(s_1,s_2) & \hdots &  sim(s_1,s_N) \\
     %       sim(s_2,s_1) & \ddots &   & \vdots \\
      %      \vdots &  & sim(s_i,s_j)  & \\
       %     sim(s_N,s_1) & \hdots &  & sim(s_N,s_N) \\
        %\end{bmatrix} 
    \]
  
    \item Repeat last step until $|R(t+1)-R(t)|>\varepsilon$.
    \item Take top ranking sentences in $\bf{R}$ for summary. 
\end{enumerate}

\subsubsection{TF/IDF}
Term Frequency(TF)/Bag of words is the count of how many times a word occurs in the given document. Inverse Document Frequency(IDF) is the number of times word occurs in complete corpus. Infrequent words through corpus will have higher weights, while weights for more frequent words will be depricated. 

Underlying steps for TF/IDF summarization are:
\begin{enumerate}
    \item Create a count vector
    \[Doc_{1} \leftarrow  <fr_{Word_1},fr_{Word_2},fr_{Word_3},..>\]
    \item Build a tf-idf matrix $W_{M\times N}$ with element $w_{i,j}$ as,
    \[w_{i,j} = tf_{i,j} \times log(\frac{N}{df_{i}})\] \\
    Here, $tf_{i,j} $ denotes term frequency of {\tt ith} word in {\tt jth} sentence, and $log(\frac{N}{df_{i}})$ represents the IDF frequency.
    
    \item Score each sentence, taking into consideration only nouns, we use NLTK POS-tagger for identifying nouns.
    \[Score(S_{o,j}) \leftarrow \frac{\sum No_{i,j}}{\sum\limits_{p=1}^N N_{p,j}} \]
    \item Applying positional weighing .
    \[Scores[S_{o,j}] = Score(S_{o,j})\times (\frac{o}{T})\]
    \[\tt{o\rightarrow Sentence\ index}\]
    \[\tt{T\rightarrow Total\ sentences\ in\ document\ j}\]
    \item Summarize using top ranking sentences.
\end{enumerate}

\subsection{Using Semantic Models}
We proceed in the same way as we did for statistical models. All the pre-processing steps remain nearly same. We can make a little change by using lemmatizer instead of stemmer. Stemming involves removing the derivational affixes/end of words by heuristic analysis in hope to achieve base form. Lemmatization, on the other hand, involves firstly POS tagging \cite{santorini1990part}, and after morphological and vocabulary analysis, reducing the word to its base form. Stemmer output for `goes' is `goe', while lemmatized output with the verb passed as POS tag is `go'. Though lemmatization may have little more time overhead as compared to stemming, it necessarily provides better base word reductions. Since WordNet \cite{pedersen2004wordnet} and Glove both require dictionary look-ups,  in order for them to work well, we need better base word mappings. Hence lemmatization is preferred.

\subsubsection{Additional Pre-processing}
\begin{enumerate}
    \item \textbf{Part of Speech(POS) Tagging:} We tag the words using NLTK POS-Tagger.
    \item \textbf{Lemmatization:} We use NTLK lemmatizer with POS tags passed as contexts.
\end{enumerate}

\subsubsection{Using WordNet}
We generated Similarity matrices in the case of Statistical Models. We will do the same here, but for sentence similarity measure we use the method devised by Dao.\cite{dao2005measuring}
The method is defined as:
\begin{enumerate}
    \item \textbf{Word Sense Disambiguation(WSD):} We use the adapted version of  Lesk algorithm\cite{lesk1986automatic}, as devised by  Dao, to derive the sense for each word.
    \item \textbf{Sentence pair Similarity:} For each pair of sentences, we create semantic similarity matrix $S$. Let $A$ and $B$ be two sentences of lengths $m$ and $n$ respectively. Then the resultant matrix $S$ will  be of size $m\times n$, with element $s_{i,j}$ denoting semantic similarity between sense/synset of word at position $i$ in sentence $A$ and sense/synset of word at position $j$ in sentence $B$, which is calculated by path length similarity using {\tt is-a} (hypernym/hyponym) hierarchies. It uses the idea that shorter the path length, higher the similarity. To calculate the path length, we proceed in following manner:-\\
   For two words $W_{1}$ and $W_{2}$, with synsets $s_1$ and $s_2$ respectively,
    \[sd(s_1,s_2) = 1/distance(s_1,s_2)\]
    \[S_{m\times n}=
        \begin{bmatrix}
            sd(s_1,s_1) &  \hdots &  sd(s_1,s_n) \\
            sd(s_2,s_1) & \ddots  & \vdots \\
            \vdots  & sd(s_i,s_j)  & \\
            sd(s_m,s_1) & \hdots &  sd(s_m,s_n) \\
        \end{bmatrix} \]
    We formulate the problem of capturing semantic similarity between sentences as the problem of computing a maximum total matching weight of a bipartite graph, where X and Y are two sets of disjoint nodes. We use the Hungarian method \cite{kuhn1955hungarian} to solve this problem. Finally we get bipartite matching matrix $B$ with entry $b_{i,j}$ denoting matching between $A[i]$ and $B[j]$. To obtain the overall similarity, we use Dice coefficient,
    \[Sim(A,B) = \frac{|A\cap B|}{|A|+|B|}\]
    with threshold set to $0.5$, and $|A|$ ,$|B|$ denoting lengths of sentence $A$ and $B$ respectively.
    \item We perform the previous step over all pairs to generate the similarity matrix $Sim^{3}_{N\times N}$.
     
\end{enumerate}
\subsubsection{Using Glove Model}
Glove Model provides us with a convenient method to represent words as vectors, using vectors representation for words, we generate vector representation for sentences. We work in the following order,
\begin{enumerate}
    \item Represent each tokenized word $w_{i}$ in its vector form \textless$ a_{i}^{1},a_{i}^{2},a_{i}^{3},\hdots a_{i}^{300}$\textgreater.
    \item Represent each sentence into vector using following equation,
    \[SVec(s_{j})=\frac{1}{|s_{j}|}\sum_{w_{i}\in s_{j}} f_{i,j}(a_{i}^{1},a_{i}^{2},\hdots a_{i}^{300})\]
    where $f_{i,j}$ being frequency of $w_{i}$ in $s_{j}$.
    \item Calculate similarity between sentences using cosine distance between two sentence vectors.
    \item Populate similarity matrix $Sim^{4}_{N\times N}$ using previous step.
\end{enumerate}

\subsubsection{Using Facebook's InferSent} 
Infersent is a state of the art supervised sentence encoding technique \cite{conneau2017supervised}. It outperformed another state-of-the-art sentence encoder SkipThought on several benchmarks, like the STS benchmark (\textit{\burl{http://ixa2.si.ehu.es/stswiki/index.php/STSbenchmark}}). The model is trained on Stanford Natural Language Inference (SNLI) dataset \cite{bowman2015large} using seven architectures  Long Short-Term Memory (LSTM), Gated Recurrent Units (GRU), forward and backward GRU with hidden states concatenated, Bi-directional LSTMs (BiLSTM) with min/max pooling, self-attentive network and (HCN's) Hierarchical convolutional networks. The network performances are task/corpus specific.\\
Steps to generate similarity matrix $Sim^{5}_{N\times N}$ are:
\begin{enumerate}
    \item Encode each sentence to generate its vector representation \textless $l_{i}^{1},l_{i}^{2},l_{i}^{3},\hdots l_{i}^{4096}$\textgreater.
    \item Calculate similarity between sentence pair using cosine distance.
    \item Populate similarity matrix $Sim^{5}_{N\times N}$ using previous step.
\end{enumerate}

\subsection{Generating Summaries}
TF-IDF scores and TextRank allows us to directly rank sentences and choose $k$ top sentences, where $k$ is how many sentences user want in the summary. On the other hand, the similarity matrix based approach is used in case of all Semantic Models, and Similarity/correlation based Statistical models. To rank sentences from Similarity matrix, we can use following approaches:-
\begin{enumerate}
    \item \textbf{Ranking through Relevance score}\\
    For each sentence $s_{i}$ in similarity matrix the Relevance Score is as:-\\
    \linebreak
        $RScore(s_{i})=\sum_{j=1}^{N} Sim[i,j]$\\
        \linebreak
    We can now choose $k$ top ranking sentences by RScores. Higher the RScore, higher the rank of sentence.
    
    \item \textbf{Hierarchical Clustering}\\
    Given a similarity matrix $Sim_{N\times N}$, let $s_{a,b}$ denote an individual element, then Hierarchical clustering is performed as follows:-
        \begin{enumerate}
            \item Initialize a empty list $R$.
            \item Choose element with highest similarity value let it be $s_{i,j}$ where,
            $i\neq j, s_{i,j}\neq0$
            \item Replace values in column and row $i$ in following manner:-\\
            \linebreak
               $s_{d,i}=\frac{s_{d,i}+s_{d,j}}{2}  ,d\in(1,N)$ \\
               \linebreak
                $s_{i,d}=\frac{s_{i,d}+s_{j,d}}{2}  ,d\in(1,N)$\\
            \item Replace entries corresponding to column and row $i$ by zeros.
            \item Add $i$ and $j$ to $R$, if they are not already there.
            \item Repeat steps 2-5 until single single non-zero element remains, for remaining non-zero element apply Step 5 and terminate.
            \item We will have rank list $R$ in the end.   
        \end{enumerate}
    We can now choose $k$ top ranking sentences from $R$.
\end{enumerate}

\subsection{Single Document Summarization}
After generating summary from a particular model, our aim is to compute summaries through overlap of different models. Let us have $g$ summaries from $g$ different models. For $p_{th}$ summarization model, let the $k$ sentences contained be:-\\
\linebreak
    $Sum_{p}\leftarrow (s_{(1,p)},s_{(2,p)}\hdots ,s_{(k,p)})$\\
   \linebreak
Now for our list of sentences $<s_1 ,s_2,s_3,.. s_n >$ we define \textbf{cWeight} as weight obtained for each sentence using $g$ models.\\
\linebreak
    $cWeight(s_{i}) = \sum_{j=1}^{g} W_{i}B(j,s_{i})$\\
    \linebreak
    Here, $B(j,s_{i})$ is a function which returns $1$ if sentence is in summary of $j_{th}$ model, otherwise zero. $W_{i}$ is weight assigned to each model without training,  
    $W_{i} = \frac{1}{g}, i\in (1,g)$

\subsection{Multi-Document/Domain-Specific Summarization}
We here use machine learning based approach to further increase the quality of our summarization technique. The elemental concept is that we use training set of $u$ domain specific documents, with gold standard/human-composed summaries, provided we fine tune our weights $W_{i} \forall i\in(1,g)$ for different models taking F1-score/F-measure.\cite{powers2011evaluation} as factor.
\[F1Score = \frac{2.precision.recall}{precision+recall}\]
We proceed in the following manner:-
\begin{enumerate}
    \item For each document in training set generate summary using each model independently, compute the $F_{1} Score$ w.r.t. gold summary.
    \item For each model, assign the weights using
        \[W_{i}=\frac{\sum_{j=1}^{v}f_{1}^{(j,i)}}{u}, i\in (1,g)\]
        Here, $f_{1}^{(j,i)}$ denotes $F_{1} Score$ for $j_{th}$ model in $i_{th}$ document.
\end{enumerate}
We now obtain \textbf{cWeight} as we did previously, and formulate cumulative summary, capturing the consensus of different models. We hence used a supervised learning algorithm to capture the mean performances of different models over the training data to fine-tune our summary.

\subsection{Domain-Specific Single Document Summarization}

As we discussed earlier, summarization models are field selective. Some models tend to perform remarkably better than others in certain fields. So, instead of assigning uniform weights to all models we can go by the following approach.
\begin{enumerate}
   \item  For each set of documents we train on, we generate document vector using bidirectional GRU ( \cite{bahdanau2014neural} as described by Zichao Yang \cite{yang2016hierarchical}for each document. We then generate complete corpus vector as follows:-
    \[cDocs=\sum_{i=1}^{v}(a_{i}^{1},a_{i}^{1},a_{i}^{1},\hdots,a_{i}^{p},)\]
    where,$v$ is total training set size, $p$ is number of features in document vector.
  \item We save  $cDocs$ and $weights$ corresponding to each corpus.
  \item For each single document summarization task, we generate given texts document vector, perform nearest vector search over all stored $cDocs$, apply weights corresponding to that corpus.
\end{enumerate}

\FloatBarrier
\subsection{Experiments}

\begin{center}
\begin{table}[p]
  \caption{Average ROUGE-2 Scores for Different Combination of Models.}
  \label{tab:parameter_estimates}
    \begin{tabular}{ | c | c | c | c | c | c | c |}
    \hline
      \multicolumn{6}{|c|}{ Models} & Score \\
      \hline

     A & B & C & D & E & F & $ROUGE2 (95\%)$\\
     \hline

    &  &   \textbullet & &   \textbullet &  &  $0.03172$  \\ \hline
        \textbullet &  &  &&      \textbullet &  &  $0.03357$  \\ \hline
        \textbullet &  &  \textbullet &  &  \textbullet &  &  $0.03384$  \\ \hline
 &      \textbullet &     \textbullet & &   \textbullet &  &  $0.03479$  \\ \hline
        \textbullet &     \textbullet &  & &        \textbullet &  &  $0.03572$  \\ \hline
        \textbullet &     \textbullet  &    \textbullet &&    \textbullet &  &  $0.03519$  \\ \hline
 &  &   \textbullet & &  &        \textbullet &   $0.03821$  \\ \hline
        \textbullet &  &  &&  &   \textbullet &   $0.03912$  \\ \hline
        \textbullet &  &  \textbullet &  &  &       \textbullet &   $0.03822$  \\ \hline
 &      \textbullet &     \textbullet & &  &        \textbullet &   $0.03986$  \\ \hline
        \textbullet &     \textbullet &  & &  &     \textbullet &   $\textbf{0.04003}$  \\ \hline
        \textbullet &     \textbullet  &    \textbullet &&  &         \textbullet &   $0.03846$  \\ \hline
 &  &   \textbullet &     \textbullet &  &  &   $0.03312$  \\ \hline
        \textbullet &  &  &       \textbullet &  &  &   $0.03339$  \\ \hline
        \textbullet &  &  \textbullet &     \textbullet &  &  &   $0.03332$  \\ \hline
 &      \textbullet &     \textbullet &     \textbullet &  &  &   $0.03532$  \\ \hline
        \textbullet &     \textbullet &  &  \textbullet &  &  &   $0.03525$  \\ \hline
        \textbullet &     \textbullet  &    \textbullet &     \textbullet &  &  &   $0.03519$  \\ \hline
 &  &   \textbullet &     \textbullet &  &  \textbullet &  $0.03721$  \\ \hline
        \textbullet &  &  &       \textbullet &  &  \textbullet &  $0.03689$  \\ \hline
        \textbullet &  &  \textbullet &     \textbullet &  &  \textbullet &  $0.03771$  \\ \hline
 &      \textbullet &     \textbullet &     \textbullet &  &  \textbullet &  $0.03812$  \\ \hline
        \textbullet &     \textbullet &  &  \textbullet &  &  \textbullet &  $0.03839$  \\ \hline
        \textbullet &     \textbullet  &    \textbullet &     \textbullet &  &  \textbullet &  $0.03782$  \\ \hline
 &  &   \textbullet & &   \textbullet &     \textbullet &  $0.03615$  \\ \hline
        \textbullet &  &  &&      \textbullet &     \textbullet &  $0.03598$  \\ \hline
        \textbullet &  &  \textbullet &  &  \textbullet &     \textbullet &  $0.03621$  \\ \hline
 &      \textbullet &     \textbullet & &   \textbullet &     \textbullet &  $0.03803$  \\ \hline
        \textbullet &     \textbullet &  & &        \textbullet &     \textbullet &  $0.03819$  \\ \hline
        \textbullet &     \textbullet  &    \textbullet &&    \textbullet &     \textbullet &  $0.03784$  \\ \hline
 &  &   \textbullet &     \textbullet &     \textbullet &  &  $0.03314$  \\ \hline
        \textbullet &  &  &       \textbullet &     \textbullet &  &  $0.03212$  \\ \hline
        \textbullet &  &  \textbullet &     \textbullet &     \textbullet &  &  $0.03426$  \\ \hline
 &      \textbullet &     \textbullet &     \textbullet &     \textbullet &  &  $0.03531$  \\ \hline
        \textbullet &     \textbullet &  &  \textbullet &     \textbullet &  &  $0.03544$  \\ \hline
        \textbullet &     \textbullet  &    \textbullet &     \textbullet &     \textbullet &  &  $0.03529$  \\ \hline
 &  &   \textbullet &     \textbullet &     \textbullet  &    \textbullet & $0.03712$  \\ \hline
        \textbullet &  &  &       \textbullet &     \textbullet  &    \textbullet & $0.03713$  \\ \hline
        \textbullet &  &  \textbullet &     \textbullet &     \textbullet  &    \textbullet & $0.03705$  \\ \hline
 &      \textbullet &     \textbullet &     \textbullet &     \textbullet  &    \textbullet & $0.03821$  \\ \hline
        \textbullet &     \textbullet &  &  \textbullet &     \textbullet  &    \textbullet & $0.03829$  \\ \hline
        \textbullet &     \textbullet  &    \textbullet &     \textbullet &     \textbullet  &    \textbullet & $0.03772$  \\ \hline
    \hline
    \end{tabular}
    \medskip

  \parbox{\linewidth}{\scriptsize%
  
  \textbf{A}$\rightarrow$ Jaccard/Cosine\ Similarity\ Matrix\\
  \textbf{B}$\rightarrow$ TextRank\\
  \textbf{C}$\rightarrow$ TFIDF\\
  \textbf{D}$\rightarrow$ WordNet\ Based\ Model\\
  \textbf{E}$\rightarrow$ Glove-vec\ Based\ Model\\
  \textbf{F}$\rightarrow$ InferSent\ Based\ Model\\
  }
  \end{table}
  \end{center}
  
  We evaluate our approaches on 2004 DUC(Document Understanding Conferences) dataset(\burl{https://duc.nist.gov/}). The Dataset has 5 Tasks in total. We work on Task 2. It (Task 2) contains 50 news documents cluster for multi-document summarization. Only 665-character summaries are provided for each cluster. For evaluation, we use ROGUE, an automatic summary evaluation metric. It was firstly used for DUC 2004 data-set. Now, it has become a benchmark for evaluation of automated summaries. ROUGE is a correlation metric for fixed-length summaries populated using n-gram co-occurrence. For comparison between model summary and to-be evaluated summary, separate scores for 1, 2, 3, and 4-gram matching are kept. We use ROUGE-2, a bi-gram based matching technique for our task.

\begin{center}
\begin{table}[H]
  \caption{Average ROUGE-2 scores for base methods.}
  \label{tab:parameter_estimates}
\begin{tabular}{ |l|c| }
  \hline
   Model & $ROUGE-2$ \\
  \hline
  Jaccard & $0.03468$ \\
  Cosine & $0.02918$ \\
  TextRank & $0.03629$ \\
  TFIDF & $0.03371$ \\
  WordNet Based Model & $0.03354$ \\
  Glove-vec Based Model & $0.03054$\\
  InferSent Based Model & $0.03812$\\
  \hline
\end{tabular}

\end{table}
\end{center}
In the \textit{Table 1}, we try different model pairs with weights trained on corpus for Task 2. We have displayed mean ROUGE-2 scores for base Models. We have calculated final scores taking into consideration all normalizations, stemming, lemmatizing and clustering techniques, and the ones providing best results were used. We generally expected WordNet, Glove based semantic models to perform better given they better capture crux of the sentence and compute similarity using the same, but instead, they performed average. This is attributed to the fact they assigned high similarity scores to not so semantically related sentences. We also observe that combinations with TF/IDF and Similarity Matrices(Jaccard/Cosine) offer nearly same results. The InferSent based Summarizer performed exceptionally well. We initially used pre-trained features to generate sentence vectors through InferSent. 

\subsection{Conclusion/Future Work}
We can see that using a mixture of Semantic and Statistical models offers an improvement over stand-alone models. Given better training data, results can be further improved. Using domain-specific labeled data can provide a further increase in performances of Glove and WordNet Models.

Some easy additions that can be worked on are:
\begin{enumerate}
    \item Unnecessary parts of the sentence can be trimmed to improve summary further.
    \item Using better algorithm to capture sentence vector through Glove Model can improve results.
    \item Query specific summarizer can be implemented with little additions.
    \item For generating summary through model overlaps, we can also try Graph-based methods or different Clustering techniques.
\end{enumerate}

\bibliographystyle{acl} 
\bibliography{research_paper_latex}

\end{document}